\documentclass[letterpaper, 10 pt, conference]{ieeeconf}
\usepackage[utf8]{inputenc}
\usepackage[table,xcdraw]{xcolor}
\usepackage[normalem]{ulem}
\usepackage{adjustbox}
\usepackage[ruled,vlined]{algorithm2e}
\usepackage{amsfonts}
\usepackage{bm}
\usepackage[labelfont=bf]{caption}
\usepackage{subcaption}
\usepackage{enumerate}
\let\labelindent\relax
\usepackage{enumitem}
\usepackage{amsmath}
\usepackage{float}
\usepackage{url}
\usepackage[super]{nth}
\usepackage{hyperref}
\usepackage{todonotes}
\usepackage{wrapfig}
\usepackage{multirow}
\usepackage{multicol}
\usepackage{tabularx,booktabs}
\usepackage{booktabs}
\useunder{\uline}{\ul}{}

\IEEEoverridecommandlockouts 
\usepackage{balance}

\IEEEoverridecommandlockouts 
\providecommand{\keywords}[1]
{
  \small	
  \textbf{\textit{Keywords---}} #1
}

\title{\LARGE \bf RASS: Risk-Aware Swarm Storage}
\author{Samuel Arseneault, David Vielfaure, Giovanni Beltrame}

\begin{document}
\maketitle

\begin{abstract}
In robotics, data acquisition often plays a key part in
unknown environment exploration. For example, storing information about the topography of the explored terrain or the natural dangers in the environment can
inform the decision-making process of the robots. 
Therefore, it is crucial to store these data safely
and to make it available quickly to the operators of the robotic
system. In a decentralized system like a swarm of robots, this
entails several challenges. To address them, we propose RASS, a
decentralized risk-aware swarm storage and routing mechanism,
which relies exclusively on local information sharing between neighbours to establish
storage and routing fitness. We test our system through thorough experiments in a physics-based simulator and test its real-world applicability with physical experiments. We obtain convincing reliability, routing speeds, and swarm storage capacity results.
\end{abstract}

\begin{keywords}
Swarm Robotics, Information Sharing, Robot Safety
\end{keywords}

\section{Introduction}
Using multi-robot systems for the exploration of unknown environments
is very appealing. Indeed, if robots do not overlap in their
exploration task, the amount of terrain covered increases
proportionally with the number of robots in the system
\cite{burgard2005coordinated}. This can, for example, be particularly
useful for search and rescue scenarios \cite{kantor2003search} where
the speed at which the environment is covered is of critical
importance. However, as the number of robots increases so does the
amount of data collected, which puts pressure on the data storage infrastructure. 
Unfortunately, multi-robot systems usually
suffer from unreliable connectivity \cite{amigoni2017multirobot}, and
directly sending the information to an external storage system
(e.g., the cloud) may not always be feasible. An alternative
to overcome this issue is using the swarm of
robots as means of distributed data storage. However, because the robots composing the swarm often have very limited memory and storage capacities, saving
large amounts of data across the swarm can prove
challenging. Furthermore, the robots in a swarm are not necessarily
reliable~\cite{winfield2006safety}: even in controlled environments, 
they are usually meant to be easily replaced. This issue
is aggravated when they must face situations that decrease their
reliability, such as exposure to dangerous environments like
search and rescue scenarios, forest fire reckoning, or nuclear
inspection and cleanup. Therefore, giving the robots a way to
eventually relay the information they acquired during their mission to
a control station for more permanent storage is a definite
advantage: it not only allows the information to be stored in a safe
location accessible by human operators, it also alleviates the memory
usage of the robots and enables continuous operation.  Nonetheless,
because in practice robots have a finite communication range, the
information collected at the periphery of the swarm usually needs to
be routed through other robots before reaching a base
station. Although forwarding the data through the shortest path
towards the base station enables prompt retrieval
\cite{dutta2020efficient}, in real-world scenarios, this approach can
be problematic for two reasons: first, such a path might not exist at
all, for example if the robots are in an area completely cut off from
external communication (e.g. a cave); second, if that path does exist,
it might be too dangerous to use. For example, a robot trying to relay
information by going through a highly radioactive area might cause a
physical failure or data corruption. Furthermore, the base station
cannot request the data, as communication with the robots is
never guaranteed. Instead, data must be autonomously forwarded to the base station
by the robots. For the same reason, and because of the overhead it
entails, it is not practical to use leaders to coordinate the storage
and routing processes. In addition, the system must take into account 
that the robots are constantly moving and acquiring new data,
meaning that storage and routing conditions are dynamic. These
constraints motivate the need for a completely decentralized,
risk-aware swarm storage system that can safely and efficiently store
data and route it towards a base station in a percolating fashion
whenever possible. In this paper, we make the following contributions
to multi-robot storage:

\begin{itemize}
    \item A distributed, dynamic and risk-aware storage system
    \item A fully decentralized risk-aware routing algorithm
\end{itemize}

We name the combination of these two contributions RASS: Risk-Aware Swarm Storage.
The rest of the article is structured as follows: in section \ref{relatedwork}, we
present related work and useful concepts upon which we build RASS; in section \ref{systemmodel} we present our system model; in section \ref{Simulations} and \ref{Physical experiments} we
detail our simulations and physical experiments along with their respective results; finally in section \ref{conclusion} we
draw some conclusions.

\section{Related Work and Background}
\label{relatedwork}

\subsection{Distributed Storage}

Distributed hash tables (DHTs) are data structures specifically
designed to partition data across a network of storage nodes in a
key-value format. They come with a set of issues susceptible of
hindering their performance, consistency and partial connectivity
being some of them~\cite{amigoni2017multirobot}. The virtual stigmergy
presented in \cite{pinciroliTuple2016} tackles the problem using
Conflict-free Replicated Data Types (CRDTs) and allowing decentralized
robust information sharing for multi-robot systems. This virtual
stigmergy is implemented directly into the Buzz programming
language~\cite{pinciroliBuzz2016}. However, in the virtual stigmergy,
each node stores a full copy of the data (in an eventually consistent
way). This means it has full redundancy and no partitioning, thus
offering good reliability but poor storage capacity optimization.

A memory-efficient storage mechanism designed for multi-robot systems
was proposed in SwarmMesh~\cite{majcherczykSwarmmesh2020}. It
uses the available memory of an agent and its connectivity with its
peers to generate unique identifiers, which are the basis of its
partitioning scheme and which represent an agent's fitness to store
data. Storage of larger items (such as binary files) in DHTs has been
studied in \cite{varadharajan2020soul}, where an auction process is
used to partition data.  Other partitioning systems have been
proposed, such as Locality-aware Distributed Hash
Tables~\cite{wu2008ldht} which posits that routing data through nodes
that are close in a network will reduce latency. Therefore, they use
information from the nodes' Autonomous System Numbers (ASNs) to
partition data.  Geographic Hash Tables~\cite{ratnasamy2002ght}, Cell
Hash Routing~\cite{araujo2005chr} and \cite{ahullo2008supporting} all
use position-based information in their partitioning systems, and the
second specifically addresses DHT implementation for ad-hoc networks
of resource-constrained nodes.

These systems show various metrics that can be used to devise a
partitioning system for a DHT. However, none of them takes into
account the effects of environmental dangers on the reliability of the
distributed storage system. RASS, our decentralized risk-aware storage and routing mechanism, aims to address this issue by including
a risk measurement in its partitioning scheme.

\subsection{Risk Assessment}

Taking risk into account when designing autonomous systems is of high
importance as excessive exposure to risk can lead to system
failures. In an unknown environment, risks are usually
tied to a certain location. Information about these locations could therefore
provide increased situational awareness for robots to effectively
perform a given mission. Belief maps and occupancy maps provide this
situational awareness by assigning values to cells of a discretized
environment. Whereas occupancy maps define the presence or absence of
a feature (e.g. fire), beliefs maps assign probabilities to the cells
of the discretized environment and usually offer better performance
\cite{stachnissMappingExplorationMobile2003}. Belief maps have been
extensively used in the past for robotic exploration
\cite{kobayashiSharingExploringInformation2002,
  kobayashiDeterminationExplorationTarget2003,
  indelmanCooperativeMultirobotBelief2018}.

A risk-aware exploration algorithm leveraging a belief map of the risk
associated with the environment has been recently proposed in
\cite{vielfaure2021dora}. In this work, robots combine efforts in building a shared belief map of the environmental risks and use it to spatially avoid the dangerous
pitfalls of the environment. The belief map is used to provide risk
awareness to an exploration algorithm which ultimately results in
fewer robot failures. In RASS, we seek to use similar shared risk awareness to establish a reliable distributed data storage and
routing mechanism, while improving its memory consumption by relying on more localized information.

\subsection{Routing}

There exist many types of routing algorithms for wireless sensor
networks. For example, packets can be explicitly routed towards a known destination
by forwarding messages towards a neighbour which optimizes a given
metric. One simple approach is to send messages through the
shortest path in terms of Euclidean distance. However, this assumes a
complete knowledge of the nodes' Cartesian coordinates, which is far
from realistic in swarm robotics. Indeed, robots might be operating
into GPS-denied environments, or they might not even be equipped with
GPS at all. A comparative study of metrics for multi-hop wireless networks shows
that hop-count performs the best in scenarios where nodes are mobile
\cite{draves2004comparison} because it is robust in dynamic topologies
\cite{watteyne2009implementation}, which is the case for RASS's
nodes. Notable implementations using this metric are given in
\cite{kuruvila2005hop,zhang2014efficient,al2019efficient}, which are
respectively greedy, power-adaptive, and grid-based routing algorithms.
Routing algorithms have also been inspired by biology: some are based
on slime molds \cite{li2011slime,jiang2018toward} and others on ant
colonies \cite{jiang2018effective,liao2008data}. However, the
former category has applications mostly in static topology networks,
which makes it ill-suited to our needs. The latter category
specifically addresses data aggregation, which is particularly
relevant to our objective our percolating data towards a base station.

\section{System Model}
\label{systemmodel}

We consider a fully decentralized multi-robot system tasked with the
exploration of a dangerous environment. The multi-robot system is denoted as the collection 
of agents $a_i \in A$. We assume the robots to have limited storage and communication 
capabilities. We also assume the presence of an operator or base station who is interested 
in collecting the data generated by the swarm. Note that this last assumption does not mean 
that our system is centralized: it is reasonable to assume that the autonomous system has 
to produce some value (i.e. data) for the humans deploying it, and the base station does 
not influence the decision-making process of the swarm, but rather act as an 
information sink. We consider the amount of data needed to be stored to be greater than the individual 
storage capabilities of the robots. It is therefore impossible for the individual agents 
to store a complete copy of the system’s data, but the data can be fully stored by the 
base station. Also, we consider that the robots are only able to communicate if they are 
within a certain communication radius $R$. As a result, if a message needs to be sent to a 
distant location (e.g. to the base station), it might need to be routed through multiple 
nodes before reaching the desired destination.

\subsection{Risk Modelling}
Radiation is known to cause performance loss and failures in robots \cite{sharp1996radiation,messenger1986effects}. We therefore adapt the risk modelling from \cite{vielfaure2021dora}, which is based on a set of independent point radiation sources $S$ with individual intensity $I_j\sim\mathcal{U}(0, 1)$. Note that we use radiation as
a \emph{model} of risk, but our method could be applied to other sources of danger:
vertical air currents, high temperature areas, etc.

Risk is assumed to be dynamic, meaning that the position of the point radiation sources $\bm{s}_j(t) \in E$ can vary across time. 
For example, radiation can spread to new areas if radioactive particles are
transported by wind. It is important to account for the dynamic nature of risk as it gives 
the system the capability to adapt to changing environments. We achieve adaptability by 
having each agent sense the radiation at every time step at its current position. The 
perceived intensity decays exponentially (with $\lambda$ as a decay parameter) as the 
Euclidean distance $\rho(\bm{x}_i)$ between $\bm{s}_j$ and $\bm{x}_i$ increases. The
total perceived radiation level by a robot $a_i$ at position $\bm{x}_i \in E$ is given by:

\begin{equation}
    r(\bm{x}_i) = b + \sum_{\bm{s_j} \in S} \frac{I_{\bm{s_j}}}{1 + \lambda\rho(\bm{x}_i)^2}
    \label{eq:radiation}
\end{equation}

and is measured by an on-board sensor with Gaussian measurement noise
$b \sim \mathcal{N}(0, 0.05)$. We posit the radiation's effect on the system is to cause 
data corruption (which is one of its possible effects~\cite{messenger1986effects}). Let the 
probability of the event of a datum $d$ getting corrupted while stored on robot $a_i$ due 
to an individual radiation  source $\bm{s}_j$ be
$\mathbb{P}(c_i = 1 | \bm{s}_j) \sim
\mathcal{B}(r_{\bm{s}_j}(\bm{x}_i))$, which follows a Bernoulli
distribution. Because we assume that the sources of radiation affect
the robots (and thus the stored data) independently, the probability of a datum being 
corrupted due to the combined effect of radiation sources follows a Bernoulli 
distribution given by:

\begin{equation}
    \mathbb{P}(c_i = 1 | S) \sim \mathcal{B}(r(\bm{x}_i)) = 1 - \prod_{\bm{s_j} \in S} 1 - \mathbb{P}(c_i = 1 | \bm{s}_j)
    \label{eq:failure}
\end{equation}

\subsection{Distributed Risk-Aware Storage}
Our risk-aware storage system is built upon three assumptions:

\begin{enumerate}
\item Because nodes exposed to a higher level of risk also have a higher
  failure probability, they should be used less, thus
  maximizing overall storage reliability
\item Efficiently moving data away from the periphery of the swarm and
  towards the base station will increase the storage capacity of the
  system and the persistence potential of the data, because the base
  station usually has more storage and reliability than the swarm
\item Percolating data from edge nodes to the base station should be
  done by choosing routes devoid of risk to minimize data loss
\end{enumerate}
From these assumptions, we derive RASS' two principal mechanisms: the routing table and the fitness-based percolation. We combine them to obtain a high-level algorithm described in Alg.~\ref{alg:rass}, in which $\mathcal{N}$ is the set of neighbours of a given agent. Because of the distributed nature of the algorithm and because all robots execute the same code (except the base station), we forego the indices notation to simplify the algorithms.

\begin{algorithm}[h]
\small
\SetAlgoLined
\DontPrintSemicolon
    \While{True}{
        $update\_routing\_table()$\;
        $update\_fitness()$\;
        \;
        \If{not $is\_fit()$ and $|\mathcal{N}| > 0$}{
            $evict\_data()$\;
        }
        \;
        $store\_measurements()$
    }
 \caption{RASS Execution Loop}
 \label{alg:rass}
\end{algorithm}

\subsubsection{Routing Table}
\label{section:routingTable}
We assume that the swarm can be represented by a
\textit{connected} graph
$\mathcal{G}$ with nodes $\mathcal{A}$ and edges $\mathcal{L}$
respectively representing agents and their wireless communication
links. In practice, this connectivity cannot be maintained at all times, because of the quality of the links in $\mathcal{L}$ and because of the movements of nodes $\mathcal{A}$. However, this assumption can be mostly maintained through mechanisms for connectivity maintenance~\cite{varadharajan2020swarm} if necessary. Furthermore, the need for this assumption can be relaxed because RASS does not need a constant link with the base station; it can opportunistically move data towards it when possible, and store them locally in the meantime.

Given the existence of at least one (multi-hop) path between any node and the base station, 
we can establish a routing table based on hop count. Because peer communication is not the 
focus of this work, the routing table held by each node only contains the minimal hop count 
between its neighbours and the base station and can therefore
be implemented with a hash table. The process to construct and periodically update this 
table involves exchanging messages between the base station and the nodes as suggested by 
\cite{abdullah2015detecting}, as we implement in Alg.~\ref{alg:routing}. Building a routing 
table with the aim of prioritizing higher capacity nodes is similar to the Gateway 
Optimization \cite{openMesh2021gateways} implemented in the B.A.T.M.A.N mesh networking 
protocol \cite{johnson2008simple}, where in our case the base station acts as a gateway.
Assuming a message can only be forwarded to an immediate neighbour within a given time 
step, the required time to build the routing table in a connected graph is bounded by 
$\Omega(1)$ and $\mathcal{O}(|A|)$ as it takes at most $|A|$ steps to send a message from 
the base station to the furthest robot in a pathological topology (a line) and $1$ step if 
the network is fully connected. However, in more realistic scenarios such as tree networks 
or scale-free topologies, building the table can be expected to take on average 
$\mathcal{O}(log|A|)$ steps because of the network's depth. We implement message forwarding through a gossip algorithm, which allows local broadcasting between robots.

\begin{algorithm}[h]
\small
\SetAlgoLined
\DontPrintSemicolon
    $routing\_table \longleftarrow listen\_neighbor\_hop\_count()$\;
    \;
    \eIf{$id = 0$}{
        $min\_hops \longleftarrow 0$
    }{
        $min\_hops \longleftarrow min(routing\_table)$\;
    }
    \;
    $broadcast(min\_hops + 1)$\;
\caption{Building/Updating the Routing Table}
\label{alg:routing}
\end{algorithm}

\subsubsection{Potential-Based Percolation}
Our risk-aware storage system draws inspiration from 
SwarmMesh~\cite{majcherczykSwarmmesh2020}, in that each node periodically assigns
itself a potential $\phi_i$ based on its fitness to store data, given by:

\begin{equation}
        \phi_i =
        \left\{ 
        \begin{array}{ll}
            \frac{1}{\alpha h_i + \beta r({\bm{x}_i})} &\text{if } m_i > 0 \\
            0 &\text{otherwise}
        \end{array} \right.
        \label{equation:fitness}
\end{equation}

 where $m_i$ is the memory available on node $i$, $r_i$ is the risk
 associated with the current node's location (stored in the
 distributed belief map) and $h_i$ refers to the minimum hop count
 required to reach the base station from $i$ as specified in the
 routing table. Parameters $\alpha$ and $\beta$ are respectively the routing weights and 
 risk weights, which allow adapting the policy based on the relative importance of the 
 routing time and the environmental, risk with respect to each other. Similarly to 
 \cite{majcherczykSwarmmesh2020}, a node  which becomes unfit to store data will evict such 
 data by moving it into its routing queue. The condition for ``unfitness'' is simply:

\begin{equation}
    T\phi_i < \max_{j \in \mathcal{N}} \phi_j
\end{equation}

where $\mathcal{N}$ and $T$ are the set of $i$'s neighbours and the fitness threshold, 
respectively. The latter's purpose is to ensure data is transferred only when the 
neighbours' max fitness is significantly higher than $\phi_i$ to avoid instability and 
overhead which would result from frequent and inefficient transfers. This fitness policy  causes data to naturally percolate along the edges towards the base station because nodes 
with a higher potential are both closer to it and
located in safer areas. When necessary, data is evicted using a Least Recently Used (LRU) policy and transmitted to the fittest neighbour.

\section{Simulations}
\label{Simulations}

\subsection{Experimental Setup}
We ran extensive simulations in a physics-based simulator, ARGoS
\cite{Pinciroli:SI2012} with models of Khepera IV
\cite{kteam2021kheperaiv} robots to eliminate the effect of potential
hardware issues on the conceptual validity of our system and also to
verify that it scales well to large swarm sizes. We executed 30 simulation
runs with 100 robots for each type of experiment to reach results
with low uncertainty. The robots are placed within a 20m by 20m arena and their communication radius $R$ is set to 3m. Three (3) radiation sources are randomly distributed in the environment around the origin. The base station is located in a corner of the arena and its storage capacity is assumed to be infinite. 

\begin{figure}[h]
	\centering
    \includegraphics[width=0.90\columnwidth]{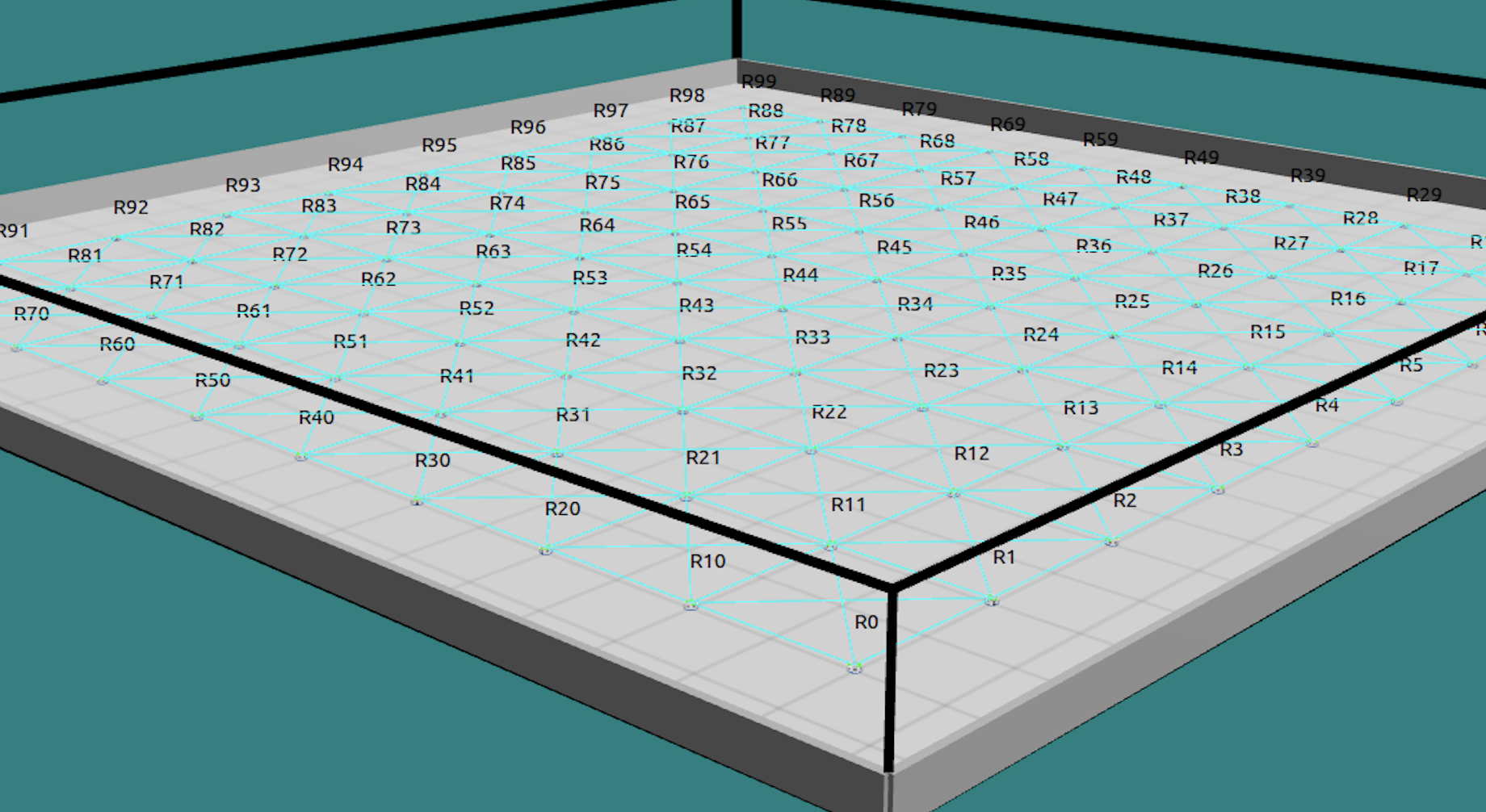}
    \caption{$400 \text{m}^2$ environment in the ARGoS simulator with 100 KheperaIV robots distributed in a grid-like pattern.}
    \label{argos}
\end{figure}

\begin{figure}[h]
	\centering
    \includegraphics[width=0.90\columnwidth]{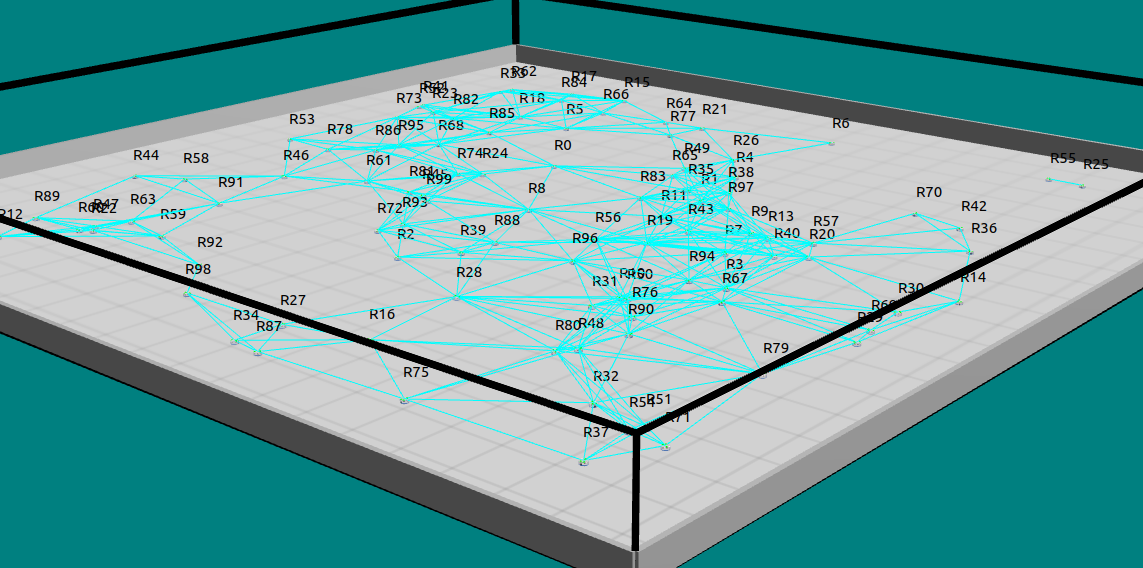}
    \caption{$400 \text{m}^2$ environment in the ARGoS simulator with 100 KheperaIV robots distributed in a scale-free pattern.}
    \label{argos}
\end{figure}

\begin{figure}[h]
	\centering
    \includegraphics[width=0.90\columnwidth]{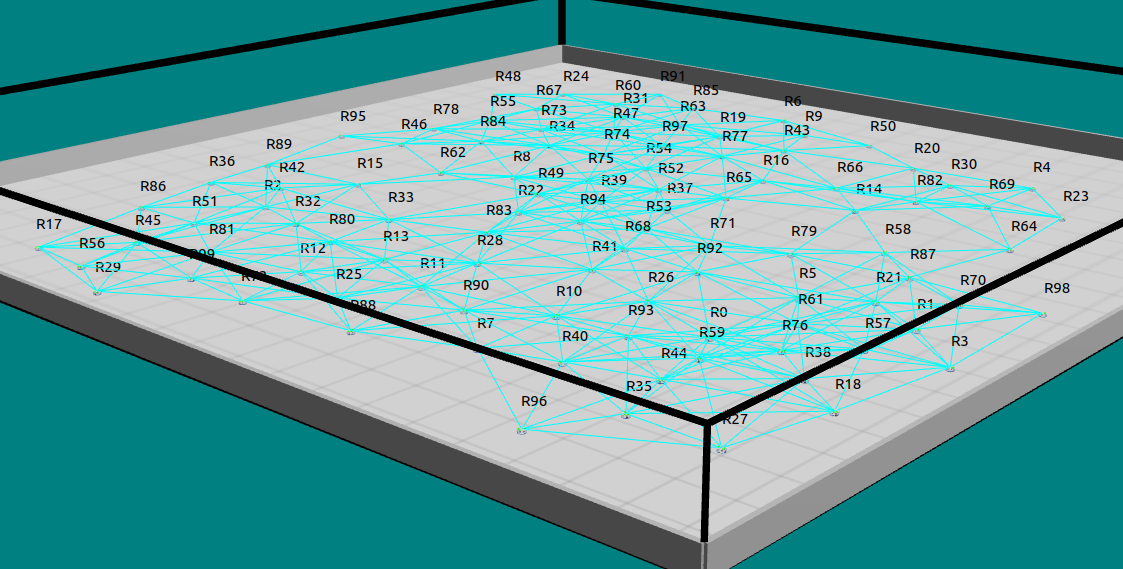}
    \caption{$400 \text{m}^2$ environment in the ARGoS simulator with 100 KheperaIV robots in a formation obtained through Lennard-Jones potential interactions.}
    \label{argos}
\end{figure}

\begin{figure}[h]
	\centering
    \includegraphics[width=0.90\columnwidth]{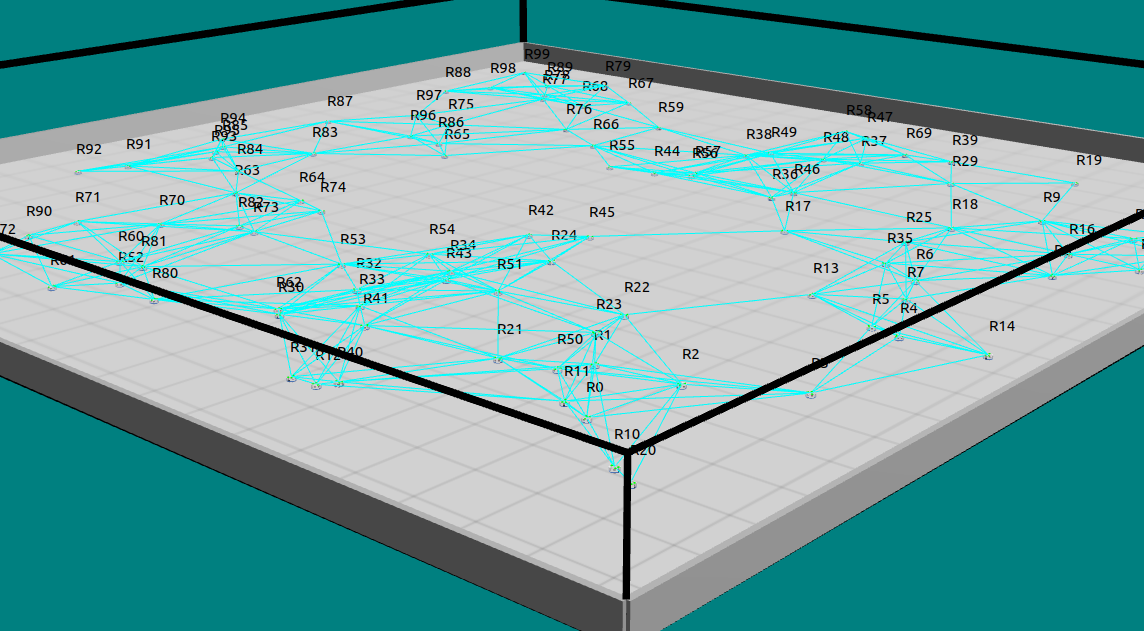}
    \caption{$400 \text{m}^2$ environment in the ARGoS simulator with 100 KheperaIV robots in a formation obtained through random walk motions.}
    \label{argos}
\end{figure}

In order to replicate realistic operation scenarios, we chose to artificially introduce bandwidth limitations. In all experiments, robots can only exchange up to 10 data items at every time step. For the same reason, our simulated robots have a limited storage capacity of 50 data items of at most 50 bytes each, as the data are simple items such as small tables or floating-point numbers. This gives a total storage capacity of 2500kb per robot. Because the robots can only exchange up to 20\% of their stored data at a given time step, it amounts to a bandwidth of 0.5kb/s. Data are generated by each robot at fixed out of phase intervals. 

To evaluate the performance of our system in different scenarios, we tested it with static topologies: a grid-like formation, and a scale-free network; as well as with dynamic topologies: a formation obtained through Lennard-Jones potential interactions and a formation evolving from random walk motions. Testing with static topologies allows us to verify applicability with fixed wireless sensor networks relevant to IoT applications, while experiments with dynamic topologies are more relevant for mobile robotics applications.

To setup RASS, we used values of $\alpha = 10$ and $beta = 1$ in Eq. \ref{equation:fitness} because risk measurement values are normalized between 0 and 1 while hop-count values are typically upper-bounded to 10 (for 100 robots). Our first benchmark algorithm is to use a fitness policy based purely on hop count, i.e. if required, data is sent only to neighbours closer to the base station. Our second comparison baseline is to store data in a virtual stigmergy (a CRDT) \cite{pinciroliTuple2016}. Using the virtual stigmergy practically ensures that no data can be lost due to corruption because it is fully replicated across the system. The first metric we used in our performance evaluation is reliability, expressed as $\frac{n_g - n_l}{n_g}$, where $n_g$ and $n_l$ are respectively the amount of data generated and lost at every time step. The second metric is the average data transfer speed, measured as the delay between the creation of a given datum and its arrival by percolation to the base station. We excluded results of this metric for the virtual stigmergy, as the stigmergy cannot include the concept of a base station (since all nodes are peers), and stigmergy propagation speeds are detailed in \cite{pinciroliTuple2016}. The third metric is the evolution of the system's total storage capacity over time, i.e. the amount of data stored by the agents and the base station combined.

\subsection{Results}

\begin{figure*}
    \centering
    \begin{subfigure}{0.30\textwidth}
        \includegraphics[width=\textwidth]{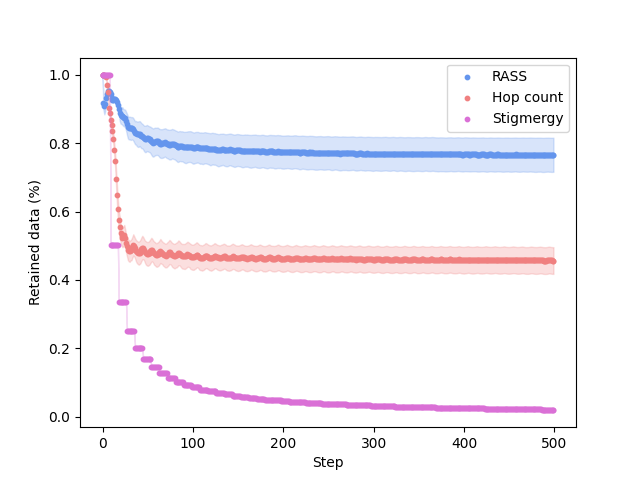}
        \caption{Evolution of reliability over time}
        \label{results:grid_100_reliability}
    \end{subfigure}
    \begin{subfigure}{0.30\textwidth}
        \includegraphics[width=\textwidth]{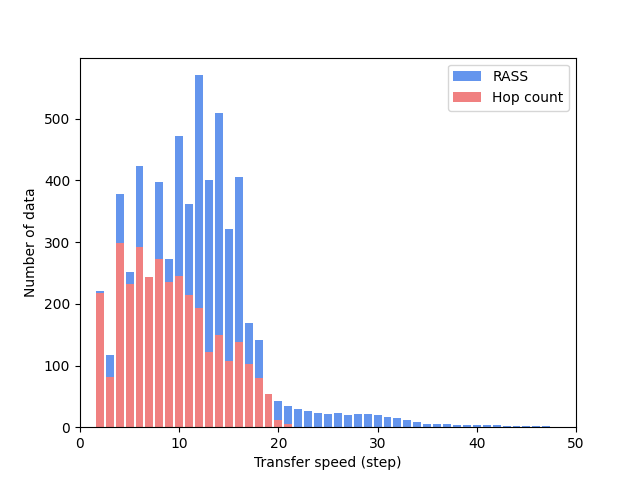}
        \caption{Distribution of transfer speeds}
        \label{results:grid_100_speed}
    \end{subfigure}
    \begin{subfigure}{0.30\textwidth}
        \includegraphics[width=\textwidth]{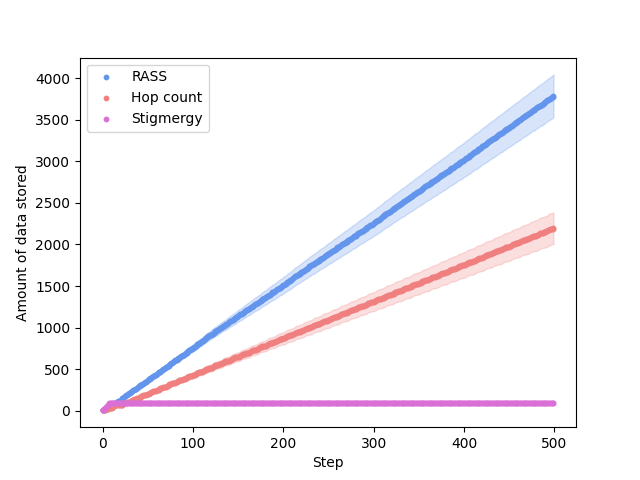}
        \caption{Evolution of total swarm storage capacity over time}
        \label{results:grid_100_storage}
    \end{subfigure}
    \caption{Performance comparison of RASS, hop count and stigmergy in a static grid-like topology.}
    \label{results:staticTopology}
    \vspace{-2mm}
\end{figure*}

\begin{figure*}
    \centering
    \begin{subfigure}{0.30\textwidth}
        \includegraphics[width=\textwidth]{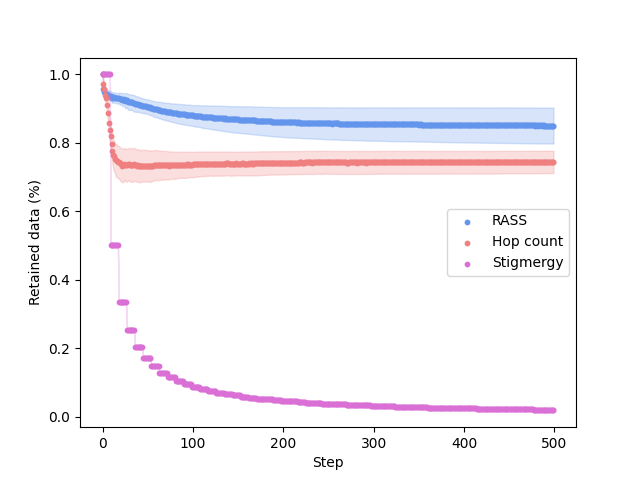}
        \caption{Evolution of reliability over time}
        \label{results:scale_100_reliability}
    \end{subfigure}
    \begin{subfigure}{0.30\textwidth}
        \includegraphics[width=\textwidth]{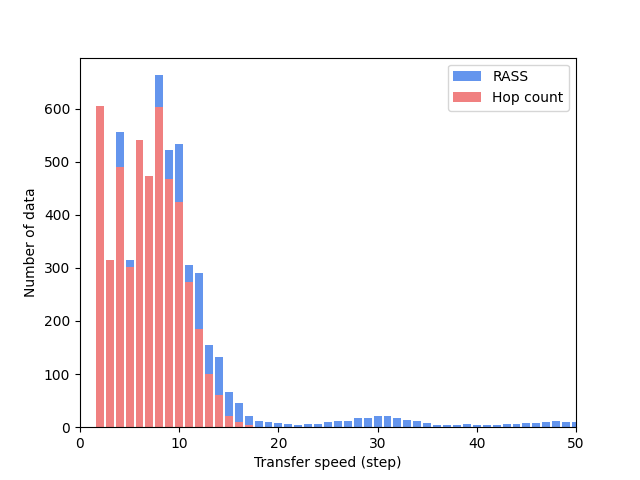}
        \caption{Distribution of transfer speeds}
        \label{results:scale_100_speed}
    \end{subfigure}
    \begin{subfigure}{0.30\textwidth}
        \includegraphics[width=\textwidth]{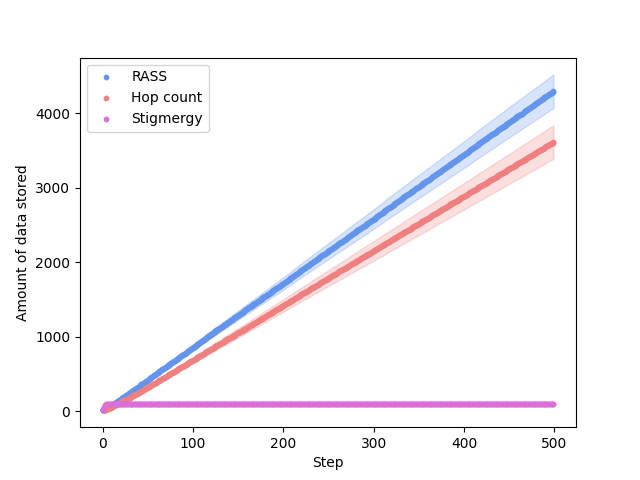}
        \caption{Evolution of total swarm storage capacity over time}
        \label{results:scale_100_storage}
    \end{subfigure}
    \caption{Performance comparison of RASS, hop count and stigmergy in a static Scale Free topology.}
    \label{results:staticTopologyScale}
    \vspace{-2mm}
\end{figure*}

\begin{figure*}
    \centering
    \begin{subfigure}{0.30\textwidth}
        \includegraphics[width=\textwidth]{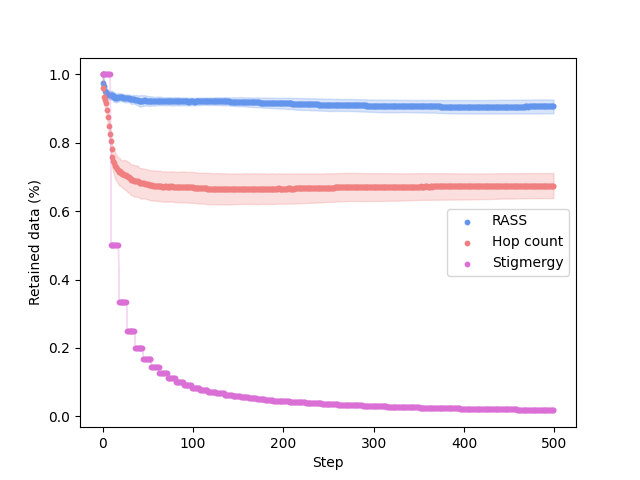}
        \caption{Evolution of reliability over time}
        \label{results:lennard_100_reliability}
    \end{subfigure}
    \begin{subfigure}{0.30\textwidth}
        \includegraphics[width=\textwidth]{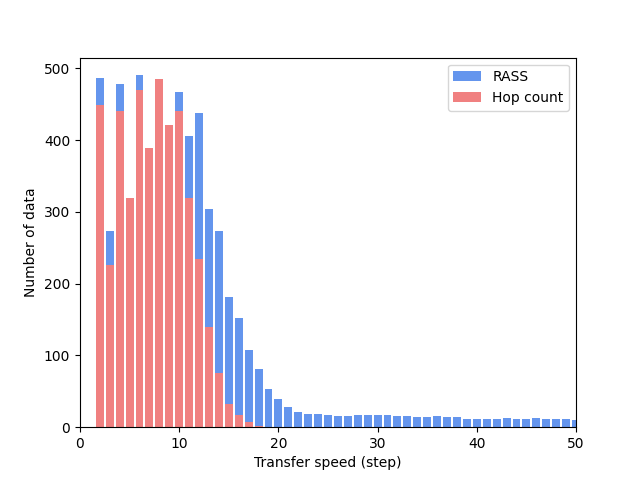}
        \caption{Distribution of transfer speeds}
        \label{results:lennard_100_speed}
    \end{subfigure}
    \begin{subfigure}{0.30\textwidth}
        \includegraphics[width=\textwidth]{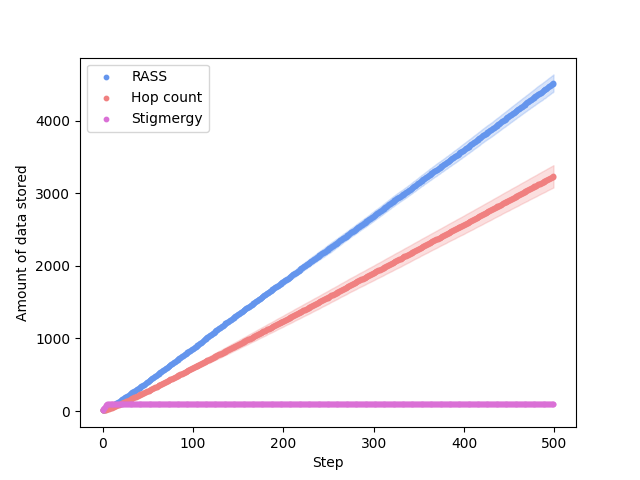}
        \caption{Evolution of total swarm storage capacity over time}
        \label{results:lennard_100_storage}
    \end{subfigure}
    \caption{Performance comparison of RASS, hop count and stigmergy in a dynamic Lennard-Jones topology.}
    \label{results:dynamicTopologyLennard}
    \vspace{-2mm}
\end{figure*}

\begin{figure*}
    \centering
    \begin{subfigure}{0.30\textwidth}
        \includegraphics[width=\textwidth]{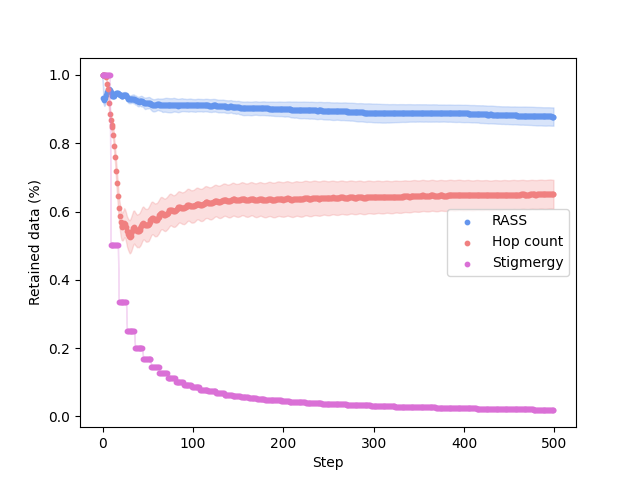}
        \caption{Evolution of reliability over time}
        \label{results:random_100_reliability}
    \end{subfigure}
    \begin{subfigure}{0.30\textwidth}
        \includegraphics[width=\textwidth]{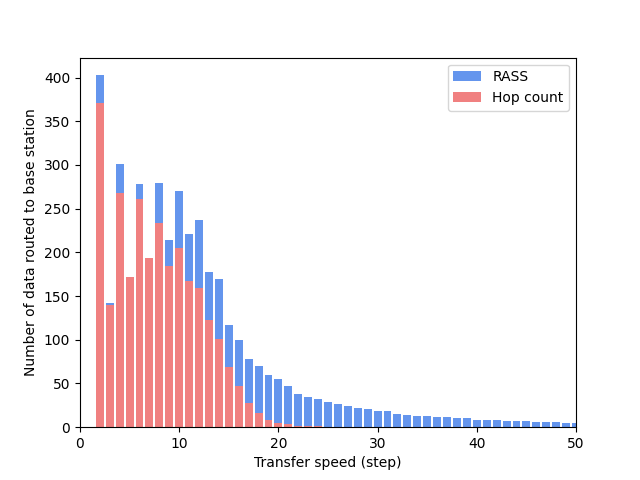}
        \caption{Distribution of transfer speeds}
        \label{results:random_100_speed}
    \end{subfigure}
    \begin{subfigure}{0.30\textwidth}
        \includegraphics[width=\textwidth]{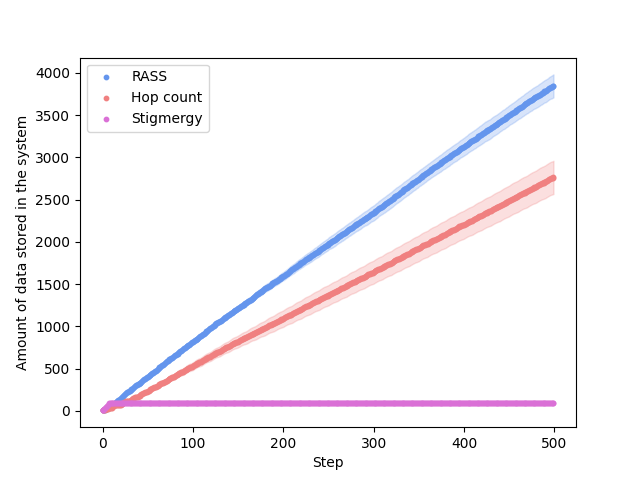}
        \caption{Evolution of total swarm storage capacity over time}
        \label{results:random_100_storage}
    \end{subfigure}
    \caption{Performance comparison of RASS, hop count and stigmergy in a dynamic random search topology.}
    \label{results:dynamicTopologyRandom}
    \vspace{-2mm}
\end{figure*}

The results obtained in the 30 simulation runs for the static topologies (grid-like and scale-free) as well as for the dynamic topologies (Lennard-Jones potential and random walk) are presented in Figs. \ref{results:staticTopology}, \ref{results:staticTopologyScale}, \ref{results:dynamicTopologyLennard} and  \ref{results:dynamicTopologyRandom} respectively.

Results show that RASS outperforms the hop-count algorithm in terms of reliability. Because 
of the risk awareness component included in its fitness policy as detailed in Eq. 
\ref{equation:fitness}, robots do not always route the data through the shortest path to 
the base station. RASS avoids the dangerous storage nodes of the system when routing data 
which explains the higher reliability levels displayed in Fig. 
\ref{results:grid_100_reliability}, Fig. \ref{results:scale_100_reliability}, Fig. 
\ref{results:lennard_100_reliability} and Fig. \ref{results:random_100_reliability} . This 
is why, on average, RASS takes 54.89\% more time to route the data to the base station when 
compared to the hop-count algorithm. The shortest path might not always be the safest one; 
RASS will take an alternate route if the risk associated with the shortest one is too high. 
On the other hand, the hop-count algorithm always takes the shortest path towards the base 
station regardless of the risk associated with it. This leads to a higher number of data 
losses due to corruption and an overall lower reliability.
However, hop count can yield faster transfer speeds as shown in Fig. \ref{results:grid_100_speed}, Fig. \ref{results:scale_100_speed}, Fig. \ref{results:lennard_100_speed}, Fig. \ref{results:random_100_speed} and table \ref{table:speed}.

\begin{table}[]
\caption{Average transfer speed and average individual memory usage with different topologies}
\label{table:speed}
\def\arraystretch{1.5}
\begin{tabular}{llcc}
\hline
\rowcolor[HTML]{C0C0C0} 
\multicolumn{1}{c}{\cellcolor[HTML]{C0C0C0}\textit{\textbf{Topology}}} &
  \multicolumn{1}{c}{\cellcolor[HTML]{C0C0C0}\textit{\textbf{Algorithm}}} &
  \textit{\textbf{\begin{tabular}[c]{@{}c@{}}Transfer speed\\ (hops)\end{tabular}}} &
  \textit{\textbf{\begin{tabular}[c]{@{}c@{}}Memory used\\ (\%)\end{tabular}}} \\ \hline
                                         & RASS      & 11.45 & 1.35   \\ \cline{2-4} 
                                         & Hop-Count & 9.11  & 0.61   \\ \cline{2-4} 
\multirow{-3}{*}{\textbf{Grid-like}}     & Stigmergy & N.A.  & 100.00 \\ \hline
                                         & RASS      & 11.44 & 1.95   \\ \cline{2-4} 
                                         & Hop-Count & 6.85  & 0.50   \\ \cline{2-4} 
\multirow{-3}{*}{\textbf{Scale-Free}}    & Stigmergy & N.A.  & 100.00 \\ \hline
                                         & RASS      & 12.51 & 1.69   \\ \cline{2-4} 
                                         & Hop-Count & 7.32  & 0.51   \\ \cline{2-4} 
\multirow{-3}{*}{\textbf{Lennard-Jones}} & Stigmergy & N.A.  & 100.00 \\ \hline
                                         & RASS      & 12.68 & 1.67   \\ \cline{2-4} 
                                         & Hop-Count & 7.76  & 0.57   \\ \cline{2-4} 
\multirow{-3}{*}{\textbf{Random Search}} & Stigmergy & N.A.  & 100.00 \\ \hline
\end{tabular}
\vspace{-3mm}
\end{table}

For the virtual stigmergy, most of the data losses can be attributed to the storage having reached its maximum capacity. Indeed, because of the fully replicated nature of the stigmergy, the memory of the agents is quickly saturated. Table \ref{table:speed} shows that on average, across the 500 steps of the simulations runs, the virtual stigmergy has 100\% of the individual memories used. This means that the nodes of the system are simply full and cannot store data anymore. In comparison, RASS uses between 1\% and 2\% of the local memories of the nodes, and hop-count is even lower at values around 0.5\%. This full redundancy prevents losing data from corruptions. However it entails a very inefficient use of the memory of the robots and ultimately leads to data losses due to insufficient memory capacity. The result is an unchanging storage capacity over time and poor reliability as shown in Fig. \ref{results:grid_100_reliability} for the virtual stigmergy strategy.

The low values of local memory usage shown in Table \ref{table:speed} for RASS and hop-count were obtained because the topologies used to test the algorithms were usually well connected in accordance with the connected graph assumption we made in \ref{section:routingTable}. For the most part, multiple routes were connecting the nodes to the base station and as a result, the collected data was routed towards the base station instead of being kept locally. Such a result implies that the system, by maintaining a low individual storage occupancy, allows the robot to adapt to situations in which they would be temporarily stranded: if their storage were to be mostly full, they would not be able to generate new data without promptly losing it. This storage buffer thus allows them to continue collecting data while being temporarily disconnected from the rest of the swarm.

\section{Physical experiments}
\label{Physical experiments}

\subsection{Experimental setup}
\begin{figure}[h]
	\centering
    \includegraphics[width=0.95\columnwidth]{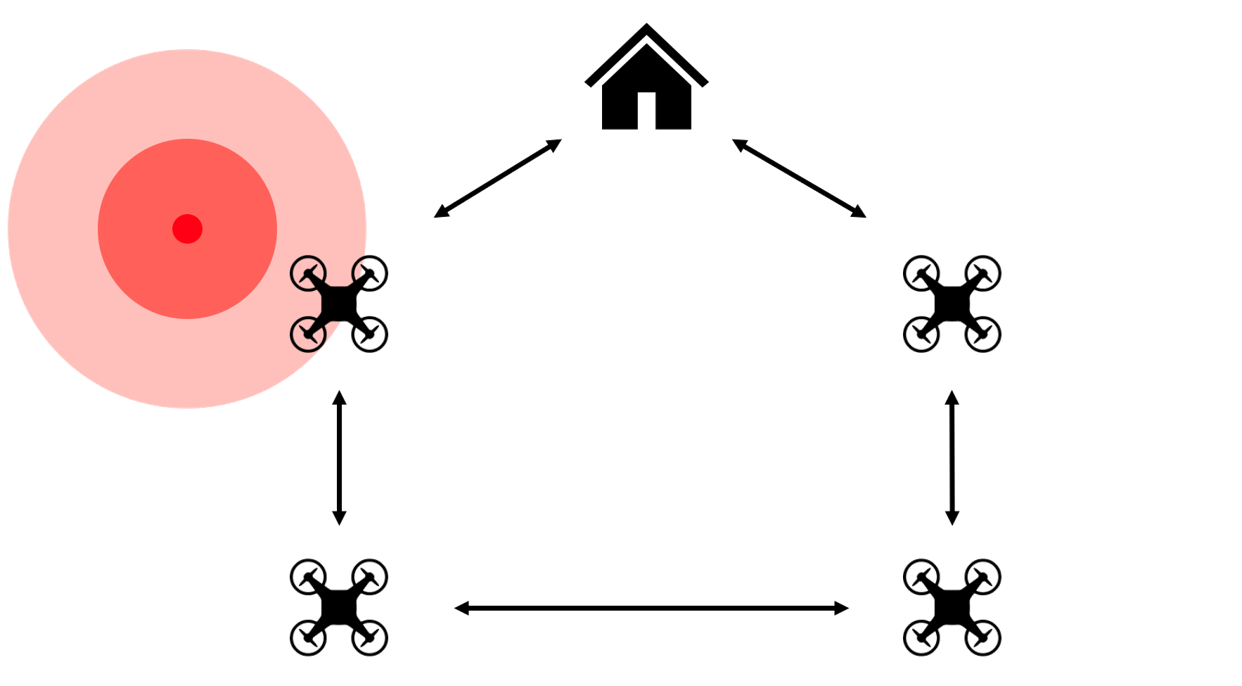}
    \caption{Topology of the 3x3m environment with 4 drones, a base station and a radiation source used in the physical experiments.}
    \label{cogniflyExperiment}
\vspace{-2mm}
\end{figure}

We evaluated RASS' performance with the same
metrics on physical robots to confirm the real-world applicability of our system. We used 5 small drones in a controlled indoor environment. These drones use standard Raspberry Pi Zeros as their main computer, meaning they have relatively low capabilities, and are therefore well suited to verify our algorithm. We designed a static topology with one drone acting as a base station and 4 others acting as agents. The radiation source (red cone) in Fig. \ref{cogniflyExperiment} is positioned to make one of the paths more dangerous, allowing us to verify if data is routed in the longer but safer path. The communication range is set to 1.5m. The topology is illustrated in Fig. \ref{cogniflyExperiment}. To assess RASS' performance, we compared it with a hop-count algorithm over 3 runs.

\subsection{Results}
\begin{figure}[h]
	\centering
    \includegraphics[width=0.95\columnwidth]{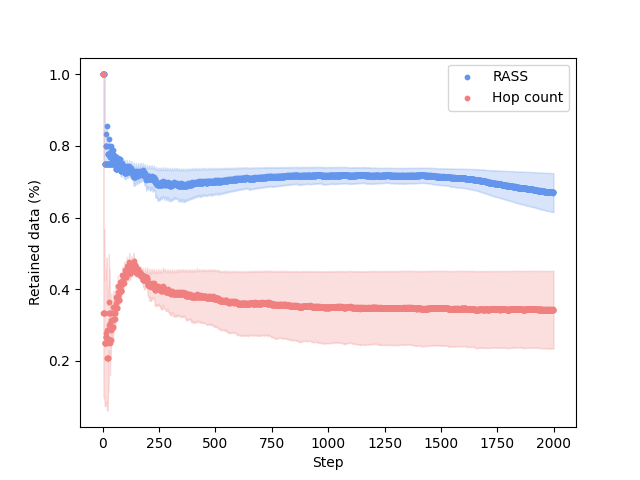}
    \caption{Evolution of reliability over time on the the physical experiments}
    \label{results:physicalRelaibility}
\vspace{-2mm}
\end{figure}

The reliability results of the physical experiments conducted on the drones are presented in Fig. \ref{results:physicalRelaibility}. They show that RASS outperforms the hop-count algorithm in terms of reliability which lead to overall greater swarm storage. Even if the topology used to assess the performance of our algorithm was simple and the number of agents in the system was limited, the physical experiments confirm the real-world applicability of RASS. Using only local interactions, it was able to choose safer paths for the data to be routed through which resulted in fewer data corruptions.

\section{Conclusions}
\label{conclusion}

We presented RASS, a Risk-Aware Swarm Storage system in which a swarm of robots can collectively store data on strategically chosen members. This choice is made without central coordination and is purely based on local information shared between the robots. This information is simply composed of risk measurements and topological distance from a robot to a base station, and used to determine a robot's fitness to store data as well as to establish the most reliable and fast route towards the base station.

We show in our experiments that RASS largely outperforms a hop-count-based solution as well as a virtual stigmergy in terms of reliability and total swarm storage capacity, while only being slightly slower in terms of percolation speed compared to the hop-count-based algorithm. RASS showed good scalability in physics-based experiments as it repeatedly performed well with a large number of robots. It performed well in experiments on physical robots.

An interesting direction for future work could be to conduct experiments in more diverse scenarios, for example in search and rescue applications where image storage and processing is required, therefore increasing the system's workload.

\bibliographystyle{IEEEtran}
\bibliography{refs.bib}

\end{document}